\documentclass[sigplan,screen]{acmart}

\usepackage{subcaption} 
\usepackage{enumitem} 

\AtBeginDocument{%
  \providecommand\BibTeX{{%
    \normalfont B\kern-0.5em{\scshape i\kern-0.25em b}\kern-0.8em\TeX}}}

\setcopyright{acmcopyright}
\copyrightyear{2020}
\acmYear{2020}
\acmDOI{10.1145/1122445.1122456}

\acmConference[]{}{}{}
\acmBooktitle{}
\acmPrice{15.00}
\acmISBN{978-1-4503-XXXX-X/18/06}

\begin{document}

\title[Detecting Tax Evasion Activities on Social Media Platforms using DNN]{Detecting Transaction-based Tax Evasion Activities on Social Media Platforms Using Multi-modal Deep Neural Networks}

\author{Lelin Zhang}
\email{lelin.zhang@uts.edu.au}
\orcid{0000-0003-0613-1362}
\affiliation{%
  \institution{University of Technology Sydney}
  \city{Sydney}
  \state{New South Wales}
  \country{Australia}
}

\author{Xi Nan}
\email{xi.nan@sydney.edu.au}
\affiliation{%
  \institution{The University of Sydney Business School}
  \city{Sydney}
  \state{New South Wales}
  \country{Australia}
}

\author{Eva Huang}
\email{eva.huang@sydney.edu.au}
\affiliation{%
  \institution{The University of Sydney Business School}
  \city{Sydney}
  \state{New South Wales}
  \country{Australia}
}

\author{Sidong Liu}
\email{sidong.liu@mq.edu.au}
\orcid{0000-0002-2371-0713}
\affiliation{%
  \institution{Macquarie University}
  \city{Sydney}
  \state{New South Wales}
  \country{Australia}
}


\begin{abstract}
Social media platforms now serve billions of users by providing convenient means of communication, content sharing and even payment between different users. Due to such convenient and anarchic nature, they have also been used rampantly to promote and conduct business activities between unregistered market participants without paying taxes. Tax authorities worldwide face difficulties in regulating these hidden economy activities by traditional regulatory means. This paper presents a machine learning based Regtech tool for international tax authorities to detect transaction-based tax evasion activities on social media platforms. To build such a tool, we collected a dataset of 58,660 Instagram posts and manually labelled 2,081 sampled posts with multiple properties related to transaction-based tax evasion activities. Based on the dataset, we developed a multi-modal deep neural network to automatically detect suspicious posts. The proposed model combines comments, hashtags and image modalities to produce the final output. As shown by our experiments, the combined model achieved an AUC of 0.808 and F1 score of 0.762, outperforming any single modality models. This tool could help tax authorities to identify audit targets in an efficient and effective manner, and combat social e-commerce tax evasion in scale.
\end{abstract}




\maketitle

\section{Introduction}
The hidden economy is perceived by tax authorities around the world, such as the Australian Taxation Office (ATO), as containing businesses that intentionally hide their income to avoid paying the right amount of taxes primarily by not recording or reporting all cash or electronic transactions; and tax evasion refers to an illegal activity that an individual or entity deliberately evades paying a true tax liability \cite{ATO2016}. According to the ATO, the hidden economy is most common in the small business segment, where about 1.6 Million small businesses operating across 233 industries are more likely to receive cash on a regular basis \cite{ATO2016}. Those caught evading taxes are usually subject to criminal charges and substantial penalties \cite{Skinner1985}.

Social media platforms such as Facebook, Twitter and Instagram serve billions of users by providing convenient means of communication, content sharing and even payment between different users, allowing for social e-commerce activities to occur. Due to their convenient and anarchic nature, it is easy for unregistered market participants to promote and conduct business activities without paying taxes. The industrial scale of these social e-commerce activities became a worldwide phenomenon, and the total transaction volume is significant. For example, Thailand was reported to have the world’s largest social e-commerce market in 2017, where 51 percent of online shoppers bought products via social media platforms, their value more than doubled to 334.2 Billion (\$10.92 Billion) \cite{PYMNTS.com2019}. According to a McKinsey report \cite{McKinseyCompany2018}, Indonesia's e-commerce spending is at least \$8 Billion a year driven by 30 Million shoppers. Among them, transactions occurred over social media platforms constituted more than \$3 Billion, accounting for around one third of the e-commerce market, projected to be \$55--65 Billion by 2022. 

Due to the concealment through text and visual content, and the social nature of social media platforms, solicited transactions are difficult for tax authorities to detect, representing the online form of transaction-based tax evasion. Tax leakage from the resulting digitalized hidden economy is significant \cite{Gaspareniene2017} and should be studied. For example, it was reported \cite{McCauley2016} that there is serious transaction-based tax evasion from cross-border online goods sales via social media platforms between China and Australia, resulting in tax losses in both countries: ``up to AUD \$1 Billion in undeclared taxable income may be slipping through the net, leaving a potential tax bill in the hundreds of millions.''

The advancement of machine learning, especially Deep Neural Networks (DNN), provides powerful tools to analyze the textual and visual content on social media. Many exciting applications have been developed with the aim to identify and combat illegal activities on social media, such as cyberbullying \cite{Agrawal2018}, counterfeit products \cite{Cheung2018}, substance use \cite{Hassanpour2019}, and drug dealing \cite{Li2019}. Compared to other applications, detecting transaction-based tax evasion activities is a unique and challenging problem. Unlike illegal substances/goods, goods sold through tax evasion transactions could be perfectly legal when the transaction happens on a registered channel. Therefore, they cannot be easily identified by simple product keywords/images. Moreover, tax evasion transactions are a mixture of different product sources and selling strategies, resulting in greater varieties of textual and visual content than most applications.

To tackle the above challenges, we built a DNN-based Regtech tool to automatically detect transaction-based tax evasion activities based on a dataset we collected from Instagram. The contributions of the tool are two-fold:

\begin{itemize}
\item We collected a dataset of 58,660 Instagram posts related to \#lipstick. We then manually labeled 2,081 sampled posts with multiple properties. The dataset provides a solid baseline to understand sales and hidden economy activities on Instagram, and facilitates the development of detection models.

\item We developed a Regtech tool to automatically detect suspicious posts of transaction-based tax evasion activities on social media platforms. The tool utilizes a multi-modal DNN model, which combines comments, hashtags and image modalities with state-of-the-art language and image networks. Our experiments shown that the combined model outperforms any single modality models, achieving an AUC of 0.808 and F1 score of 0.762. The tool could greatly increase the effective detection rate and save enforcement costs for tax authorities.
\end{itemize}

\section{Related Work}
It is difficult for governments to detect transactions in the hidden economy. Associated tax evasion has been a long-standing topic studied by tax administration and compliance scholars \cite{Rogoff2017}. Since transaction-based tax evasion activities moved onto social media platforms, traditional tax auditing detection methods failed to be implemented effectively by tax authorities, and little research has been done. The main focus of tax authorities on social media platforms is to detect taxpayers who evade income tax. They aim to catch taxpayers directly, that is, they seek to detect whether the taxpayer declared their income sourced from their social media accounts honestly \cite{Williamson2016}.  Each taxpayer is one detection point.  In Australia, the ATO has hired a team of data mining experts to look at things like Facebook and Instagram to see if income reported by taxpayers match up with their actual revenue \cite{Williamson2016}.

Transaction-based tax evasion differs from income tax evasion, taxes being evaded are transaction taxes, such as the GST, and each transaction is a stand-alone detection point, giving rise to many more detection points than the number of taxpayers. The Uganda government has taken advantage of technology and is collecting a social media tax based on the taxpayer’s daily use of the platform, but failing to take into account transaction activities\cite{Ratcliffe2019}.

As the social media becomes increasingly multi-modal and the unregistered selling activities become more sophisticated, it is essential to assess both textual (e.g., hashtags and comments) and visual (e.g., image and video) content to decide if a post on the social media has the intention to facilitate tax evasion transactions. A successful automatic detection system needs to handle both textual and visual information in a robust and efficient manner. Thankfully, with the recent development of DNN methods, we have seen exciting breakthroughs in many computer vision (CV) and natural language processing (NLP) tasks. In CV, from AlexNet \cite{Krizhevsky2012}, GoogLeNet \cite{Szegedy2015}, ResNet \cite{He2016} to EfficientNet \cite{Tan2019}, deep convolutional neural networks (DCNN) are becoming more powerful in terms of accuracy, scalability and efficiency. Similarly, the NLP models evolved from word embedding (Word2Vec \cite{Mikolov2013}), contextual word embedding (ELMo \cite{Peters2018}) to transformers (BERT \cite{Devlin2018}, XLNet \cite{Yang2019}), increasing in sophistication. 

Many existing works analyze textual and visual contents on social media platforms using DNN models. In \cite{Agrawal2018}, the authors proposed to detect cyberbullying on Instagram using a variety of textual and visual features, including Word2Vec features from comments, and DCNN features from images. In \cite{Cheung2018}, a DCNN model is used to discover counterfeit sellers on two social media platforms based on shared images. In \cite{Hassanpour2019}, DCNN for images and long short-term memory (LSTM) \cite{Hochreiter1997} for text are used to extract predictive features from Instagram posts to access substance use risk. In \cite{Li2019}, LSTM models are used to extract features from comments and hashtags, the dual-modal features are then combined to detect illicit drug dealing activities. In \cite{Zhang2018}, the authors proposed a framework to predict post popularity using both image and text data from the posts by a user. 

The majority of existing works process contents of different modalities with separate NLP/CV models, then combine the features using concatenating, stacking or embedding layers to form an end-to-end DNN model. We developed our multi-modal DNN model following the same idea with recent NLP/CV models: adapter-BERT \cite{Houlsby2019} for hashtags and comments, and EfficientNet \cite{Tan2019} for images. The aim is to have a modularized structure with the flexibility to change processing models for each modality, and the extensibility to incorporate more modalities, while keeping the whole network trainable from end-to-end. 

\section{The Instagram Dataset}

\begin{figure*}
     \centering
     \begin{subfigure}[t]{0.45\textwidth}
         \centering
         \includegraphics[width=\textwidth]{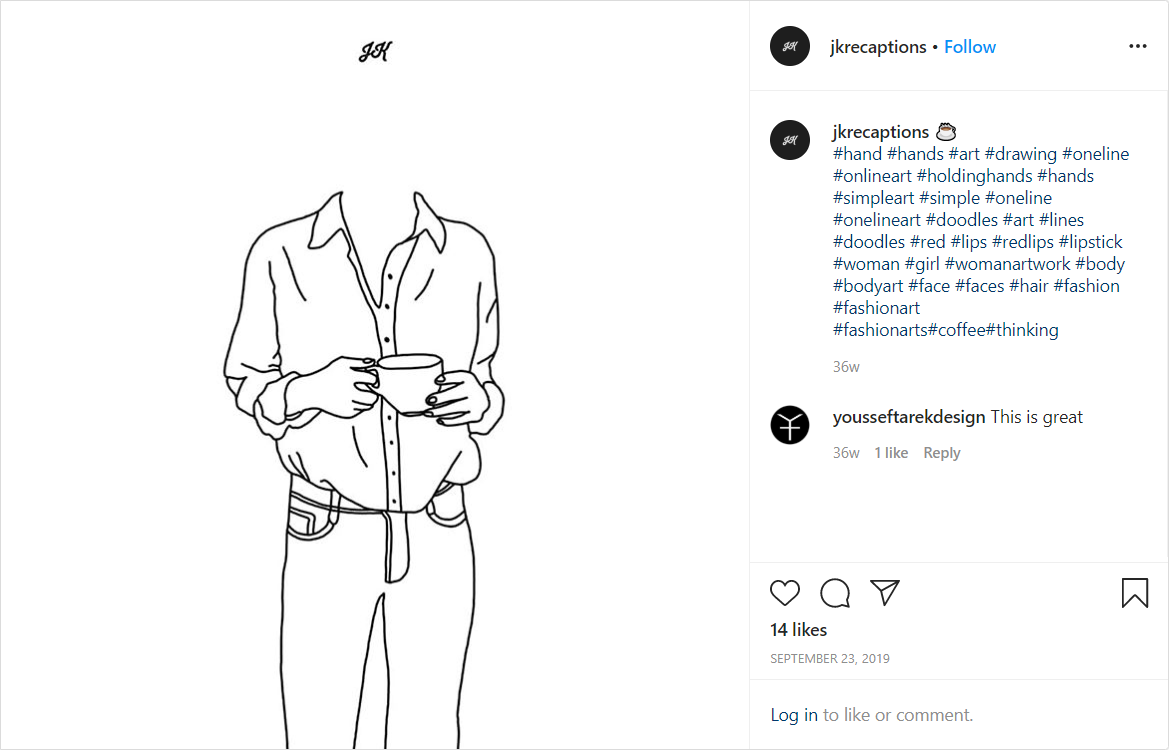}
         \caption{A post (\url{https://www.instagram.com/p/B2uFLpmnXLZ/}) not relevant to \#lipstick with no intention to sell. This post is not related to tax evasion activities.}
         \label{fig:posts:a}
     \end{subfigure}
     \hfill
     \begin{subfigure}[t]{0.45\textwidth}
         \centering
         \includegraphics[width=\textwidth]{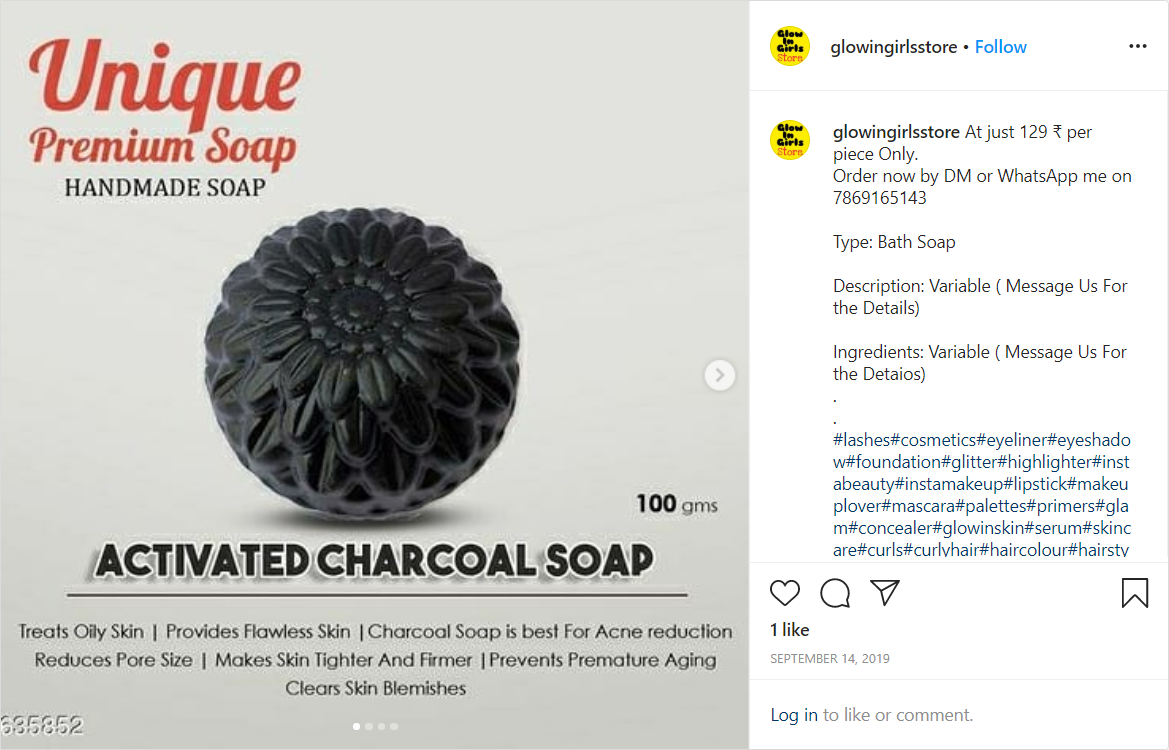}
         \caption{A post (\url{https://www.instagram.com/p/B2Y9FItjKxZ/}) not relevant to \#lipstick, but a Daigou is trying to sell a soap product. This post is related to tax evasion activities as the poster is using Instagram to generate unregistered transactions.}
         \label{fig:posts:b}
     \end{subfigure}
     
     \bigskip
     \begin{subfigure}[t]{0.45\textwidth}
         \centering
         \includegraphics[width=\textwidth]{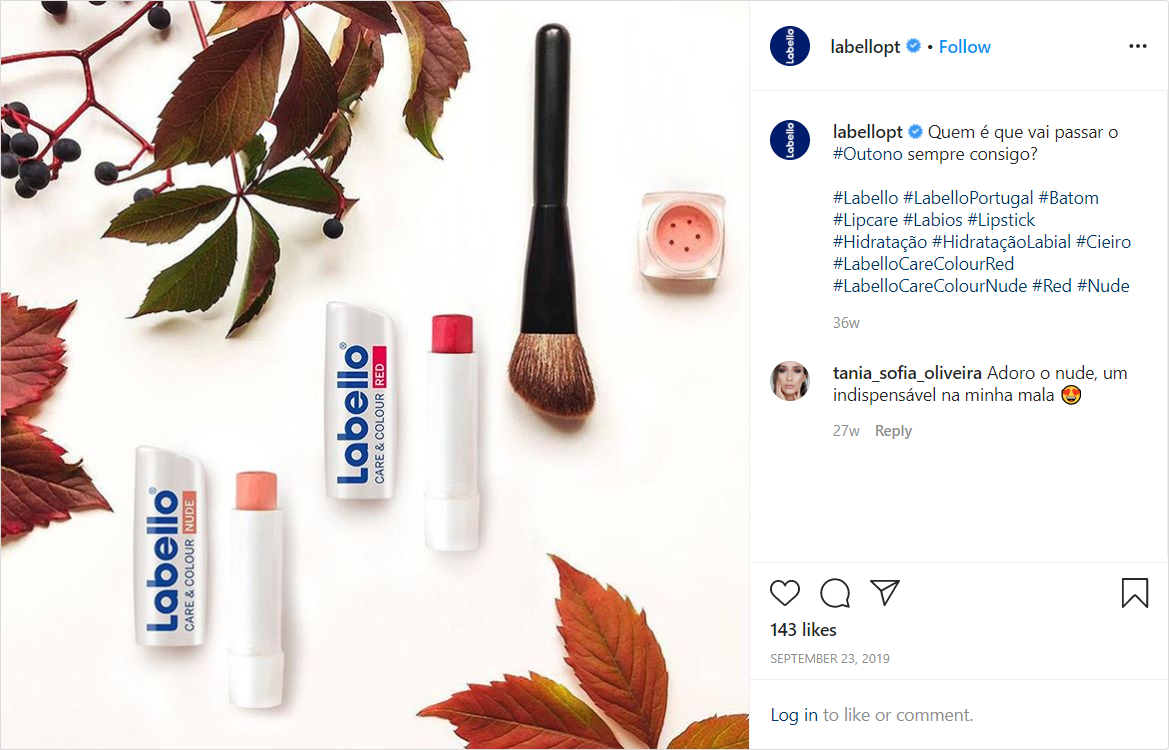}
         \caption{A post (\url{https://www.instagram.com/p/B2v3pAIgD6q/}) relevant to \#lipstick, where a brand is advertising its lipstick. This post is not related to tax evasion activities as the poster is promoting a registered channel (brand owner).}
         \label{fig:posts:c}
     \end{subfigure}
     \hfill
     \begin{subfigure}[t]{0.45\textwidth}
         \centering
         \includegraphics[width=\textwidth]{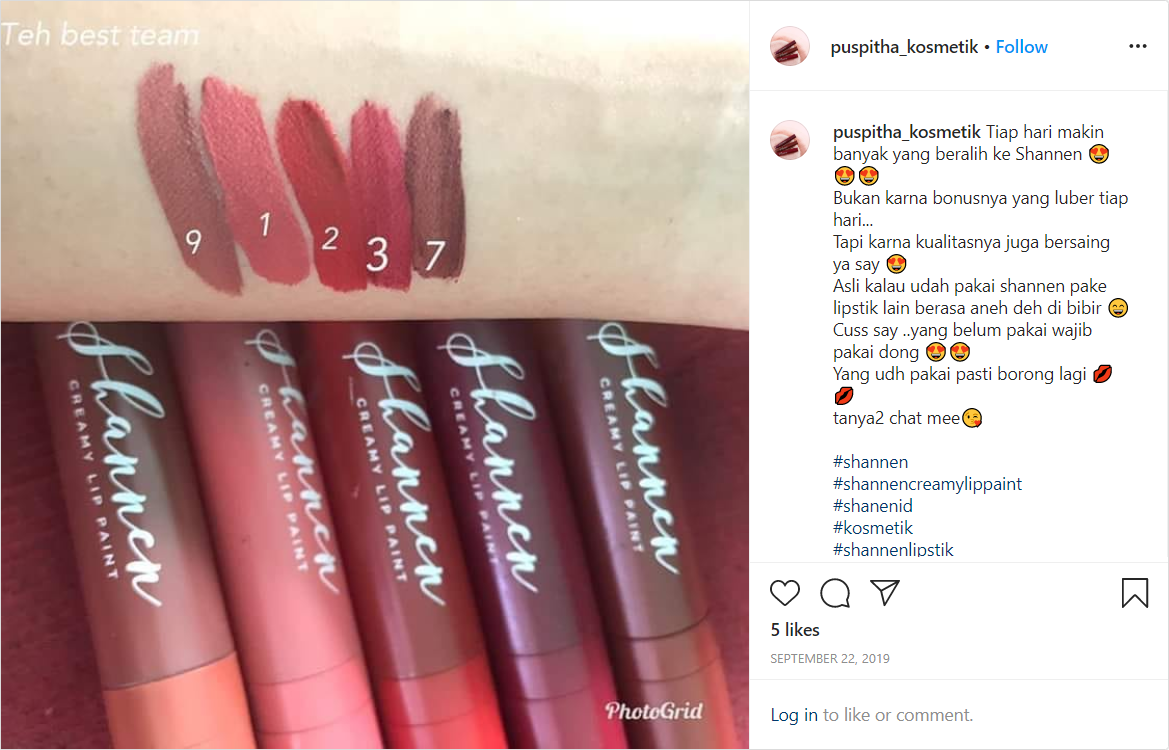}
         \caption{A post (\url{https://www.instagram.com/p/B2snt6tAVsZ/}) relevant to \#lipstick, where an unregistered producer is posting advertisement of lipsticks. This post is related to tax evasion activities as the poster is using Instagram to generate unregistered transactions.}
         \label{fig:posts:d}
     \end{subfigure}
     
     \caption{Example posts from the dataset}
     \label{fig:posts}
\end{figure*}
\subsection{Data Collection}

To build our dataset, we crawled publicly available posts and their corresponding poster information from Instagram. In the proof-of-concept stage, the posts are collected using hashtag \#lipstick during 22$^{nd}$ to 26$^{th}$ September, 2019. For posts, we collected the username, post timestamp, number of likes, image, post text, and all its comments (we include the original post text as the first comment due to the way Instagram presents the posts). We also extracted the hashtags from the comments as they usually form a significant part of the textual information. Moreover, although not used at this stage, we also collected posters’ user information, including the username, number of followers, number of following users, number of posts, and the user bio. 

We collected a total of 58,660 posts (short-lived and duplicated posts included), then randomly sampled 3,000 posts for manual data labelling. As some of the Instagram posts were deleted after a short period of time, we check the availability of the posts in November 2019 to ensure all the labelled posts are available for more than 50 days. 711 posts have been deleted at the time of checking. Moreover, we also removed 148 duplicated posts due to the way they are displayed on Instagram. This produced a dataset of 2,081 unique posts sampled from 22$^{nd}$ to 26$^{th}$ September, 2019. 

\subsection{Data Labelling}
For each collected post, we labeled according to 9 properties (as shown in Table~\ref{tab:labels}): availability of post; its relevance to the search keyword \#lipstick; its selling intention; its source; its relation to hidden economy; its image type; the language of the text; the existence of other contact details; and the type of other contact details left on the post.

For transaction-based tax evasion, the key property is the post's relationship to hidden economy, which is defined as posts by unregistered sellers and producers who intend to generate hidden economy sales. During the process, Instagram is either their main source of customers, or it is a useful tool to direct sales to the account holder’s other social media tool for the completion of the sale.

\newcommand{\labelColWidth}{0.15\textwidth}
\newcommand{\descColWidth}{0.3\textwidth}
\newcommand{\optColWidth}{0.5\textwidth}
\begin{table*}
  \caption{The 9 data labels and their corresponding post properties}
  \label{tab:labels}
  \begin{tabular}{p{\labelColWidth} p{\descColWidth} l}
    \toprule
    Label & Description & Options\\
    \midrule
      Availability
      &
      Whether the post is still available (not being deleted) at time of labelling
      &
      \begin{minipage}[t]{\optColWidth}
      \begin{itemize}[leftmargin=*]
      \item Y: the post is still available
      \item N: the post is not available
      \end{itemize}
      \end{minipage}
    \\
    \midrule
      Relevance
      &
      Whether the post is relevant to \#lipstick
      &
      \begin{minipage}[t]{\optColWidth}
      \begin{itemize}[leftmargin=*]
      \item Y: the post is relevant to \#lipstick or lipstick-related items (Figures~\ref{fig:posts:c} and \ref{fig:posts:d})
      \item N: the post is not relevant to \#lipstick and it shows something else (Figures~\ref{fig:posts:a} and \ref{fig:posts:b}) 
      \end{itemize}
      \end{minipage}
    \\
    \midrule
      Selling intention
      &
      Judging from the text and image, does the poster have an intention to sell or will the post lead to a potential sale?
      &
      \begin{minipage}[t]{\optColWidth}
      \begin{itemize}[leftmargin=*]
      \item Y: Yes, the poster has an intention to sell or the post will lead to a potential sale (Figures~\ref{fig:posts:b}, \ref{fig:posts:c} and \ref{fig:posts:d})
      \item N: No, the poster does not have intention to sell or the post will not lead to a potential sale (Figure~\ref{fig:posts:a})
      \end{itemize}
      \end{minipage}
    \\
    \midrule
      Source of the post
      &
      Judging from the text and image of each post, what is the nature of the poster?
      &
      \begin{minipage}[t]{\optColWidth}
      \begin{itemize}[leftmargin=*]
      \item I: Individuals who share posts on Instagram and the account for that individual is of personal nature and is used purely for private sharing purposes (Figure~\ref{fig:posts:a})
      \item S: Cosmetics retail shops that have created their own Instagram accounts for advertising and promotion purposes
      \item B: Brands such as Dior or Mac operating their own official accounts on Instagram for advertising purposes (Figures~\ref{fig:posts:c})
      \item M: Makeup artists/makeup tutors/models who recommend products to their fans on Instagram, or to show off how the lipstick looks in a full makeup to promote products or services
      \item D: Daigous (purchasing agents) who purchase lipsticks on behalf of their customers (Figure~\ref{fig:posts:b}).
      \item P: Unregistered producers who produce lipsticks without license and sell their homemade products (Figure~\ref{fig:posts:d}) 
      \end{itemize}
      \end{minipage}
    \\
    \midrule
      Hidden economy transactions
      &
      Judging from the text and image, is the post related to hidden economy transactions and thereby resulting in tax evasion?
      &
      \begin{minipage}[t]{\optColWidth}
      \begin{itemize}[leftmargin=*]
      \item Y: Yes, the post is related to hidden economy transactions and will result in tax evasion (Figures~\ref{fig:posts:b} and \ref{fig:posts:d})
      \item N: No, the post is not related to hidden economy transactions and will not result in tax evasion (Figures~\ref{fig:posts:a} and \ref{fig:posts:c})
      \end{itemize}
      \end{minipage}
    \\
    \midrule
      Image content type
      &
      What is the content of the image on the post?  
      &
      \begin{minipage}[t]{\optColWidth}
      \begin{itemize}[leftmargin=*]
      \item B: A body part, such as the lip or arm
      \item P: A product, such as matte lipstick, glossy lipstick or balm tint (Figures~\ref{fig:posts:b} and \ref{fig:posts:c})
      \item P+B: Both the product and body part (Figure~\ref{fig:posts:d})
      \item A: An advertisement or marketing poster
      \item S: A screenshot of the conversation between two users
      \end{itemize}
      \end{minipage}
    \\
    \midrule
      Language
      &
      What language does the poster write in?
      &
      \begin{minipage}[t]{\optColWidth}
      E.g., English, Indonesian, Malaysian, Thai, Spanish, Chinese, French, Arabic and Vietnamese
      \end{minipage}
    \\
    \midrule
      Has other contact details
      &
      Were other contact details left on the post (most posters who aim to make sales will leave contact details of other social media platforms/messaging apps)?
      &
      \begin{minipage}[t]{\optColWidth}
      \begin{itemize}[leftmargin=*]
      \item Y: Yes, there are other contact details left on the post (Figure~\ref{fig:posts:b})
      \item N: No, there are no other contact details left on the post (Figures~\ref{fig:posts:a}, \ref{fig:posts:c} and \ref{fig:posts:d})
      \end{itemize}
      \end{minipage}
    \\
    \midrule
      Specifics of other contact details
      &
      What are the other contact details left on the post (if any)? 
      &
      \begin{minipage}[t]{\optColWidth}
      E.g., WhatsApp, WeChat, Facebook, Line, Phone, etc.
      \end{minipage}
    \\
  \bottomrule
\end{tabular}
\end{table*}

\section{Deep Neural Network Models}

\begin{figure*}
  \centering
  \includegraphics[width=\textwidth]{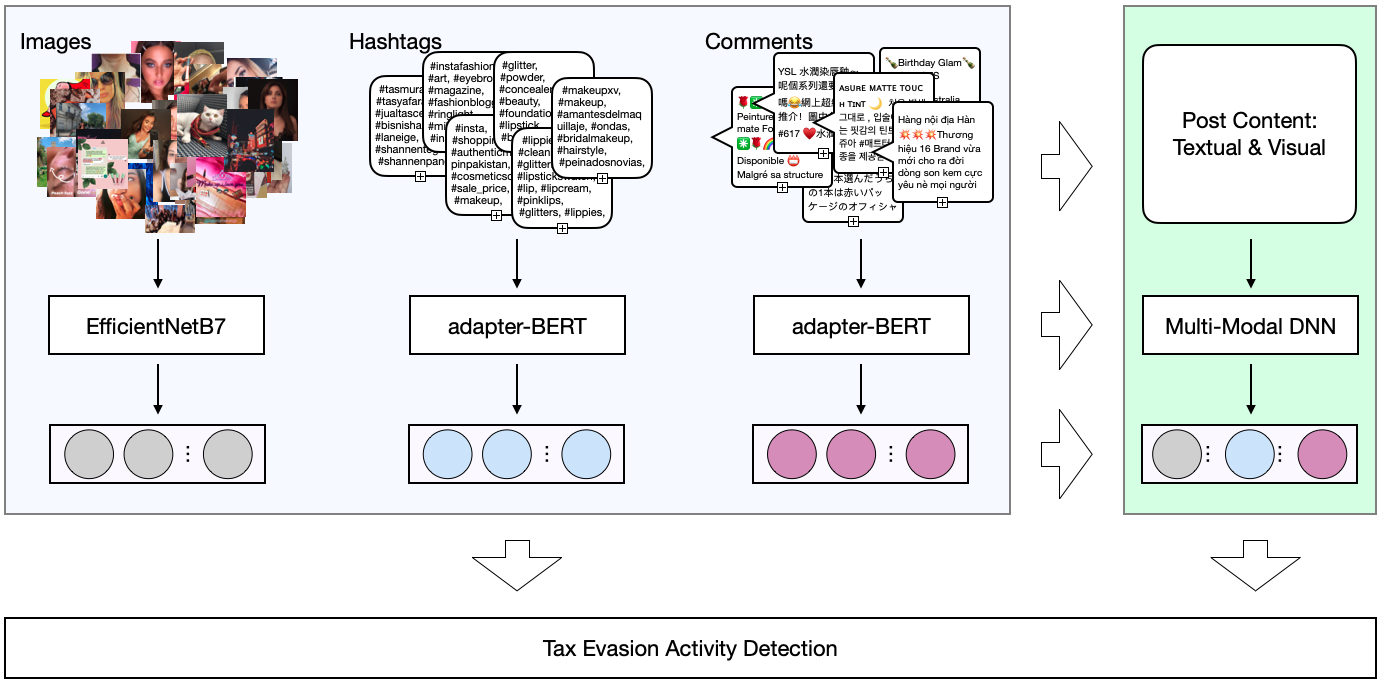}
  \caption{The schematic flow-chart of the proposed method}
  \label{fig:flowchart}
\end{figure*}

To detect transaction-based tax evasion activities of Instagram users, we analyzed their posts to extract features from the posted images, hashtags, and comments. As images and texts have different data structures and modalities, two DNN architectures, i.e, adapter-BERT \cite{Houlsby2019} and EfficientNet \cite{Tan2019}, were used for text and image feature extraction respectively. The features were extracted automatically from the post data using the established DNN models and were mapped into a joint feature space of both image and text features. A multi-modal DNN model was then developed, which took the joint feature as input, to detect transaction-based tax evasion activities. The workflow of the proposed method is illustrated in Figure~\ref{fig:flowchart}. 

\subsection{Image and Text Feature Extraction}
The adapter-BERT architecture \cite{Houlsby2019} was used to extract features from the textual content, i.e., hashtags and comments. Since the textual content contains multiple languages, a multilingual tokenizer was used to convert words into token vectors, which supported all the languages in our dataset as presented in Table 1. For image feature extraction, the EfficientNetB7 model \cite{Tan2019} was used, as it represents the state-of-the-art in object detection while being 8.4 times smaller and 6.1 times faster on inference than the best existing models in the ImageNet Challenge \cite{Deng2009}. The image and text features were then concatenated, resulting in a high dimensional feature vector for each post sample. The text features extracted by adapter-BERT are 768-dimensional for both the hashtag input and the comment input, and the image features extracted by EfficientNetB7 are 2,560-dimensional, so the joint feature space is 768 + 768 + 2560 = 4096-dimensional.

\subsection{Multi-Modal DNN Model}

We implemented a multi-modal DNN model by combining three different basic models, including two adapter-BERT models for hashtags and comments, and an EfficientNetB7 model for images. The logit layers of the three models were merged using a concatenate layer. A dropout layer was then added on top of the concatenate layer, followed by an output dense layer, where a positive prediction corresponds to a tax evasion activity, vice versa. 

Both adapter-BERT models were implemented using the bert-for-tf2 python package (v0.14) \cite{kpe2020} and pre-trained using the entire Wikipedia dump for the top 100 languages in Wikipedia \cite{Devlin2020}. The same meta-parameters were used for building the adapter-BERT models: max sequence length = 64, adapter size = 64. The EfficientNetB7 model was implemented using the efficientnet package (v1.10) \cite{Yakubovskiy2020}. The ImageNet pre-trained weights were used to initialize the EfficientNetB7 model.  

To address the imbalanced distribution of the positive and negative samples, we set the class weights to (negative: 0.4, positive: 1.6) according to the ratio of each class sample size to total sample size scaled by the number of classes. The Adam optimizer \cite{Kingma2014} with a learning rate of 0.0001 was used for training the model in 100 epochs.  

\section{Results}

400 posts were randomly selected for testing the model, and the remaining 1,681 posts were used for training. An internal validation set (20\% of the training samples) was split from the training set and used to evaluate the model’s performance during training. To test the contribution of individual text and image inputs and the effectiveness of multi-modal inputs, we compared our proposed model with the three basic DNN models using only hashtag, comment and image inputs respectively. The same initialization method, meta parameter, optimization method and learning rate as for the multi-modal model were used for the three basic models. We used Precision, Recall, F1 Score, and the Area Under Curve (AUC) to evaluate the models’ performance. 

\begin{table}
  \caption{Performance of the DNN models based on different input features}
  \label{tab:results}
  \begin{tabular}{lrrr}
    \toprule
    Input & Precision & Recall & F1 Score\\
    \midrule
    hashtags & 0.444 & \textbf{0.890} & 0.593\\
    comments & 0.656 & 0.855 & 0.742\\
    images & \textbf{0.756} & 0.645 & 0.696\\
    multi-modal & 0.722 & 0.807 & \textbf{0.762}\\
  \bottomrule
\end{tabular}
\end{table}

\begin{figure}
  \centering
  \includegraphics[width=\linewidth]{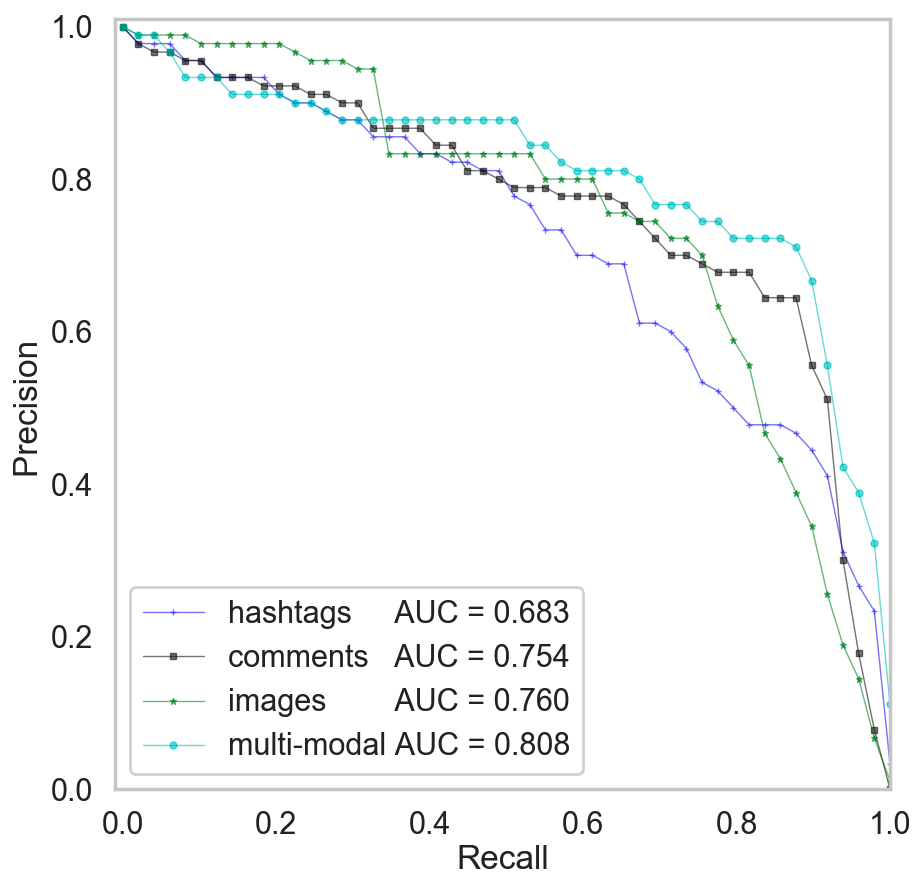}
  \caption{ROC curves of the DNN models based on different input features}
  \label{fig:roc}
\end{figure}

The results of the proposed multi-modal DNN model and the compared individual basic models are presented in Table~\ref{tab:results}, and the Receiver Operating Characteristics (ROC) curves of these models are illustrated in Figure~\ref{fig:roc}. The model using hashtag features has the most imbalanced performance with the highest recall (0.89), but the lowest precision (0.444) and F1 score (0.593). The model using comment features is less imbalanced and has a substantially better F1 score (0.742).  The model using image features achieved the highest precision (0.756), showing that it was most sensitive to tax evasion activities, but its overall performance (F1 score = 0.696) was compromised with the lowest recall (0.645). The differences in precision and recall between the image model and text models indicate that the visual and textual contents may provide complementary information to each other for hidden economy activity detection. This finding is further evidenced by the improved performance of the multi-modal model (F1 score = 0.762, AUC = 0.808), which outperformed any single modality models. 

\section{Discussion}

In order for tax authorities to effectively deploy this technology to aid in detecting transaction-based tax evasion, cost is an important factor. The very few examples of actual implementation of data-matching by tax authorities have been somewhat cost ineffective. Take the work-program of the ATO as example, in order to mitigate the major tax integrity risk of the hidden economy, their budget was AUD 39.5 Million in 2015-16, employing around 400 people, of which only 6 were data mining experts \cite{ATO2016}. Their work includes manually viewing social media posts.

In terms of cost, the proof-of-concept stage of our model was time and labor efficient, where 3 labelers spent 60 hours to complete manual labelling. Our model markedly improves the efficiency at detecting and confirming posts that relate to transaction-based tax evasion. Without the detection model, tax officers will need to randomly select the posts, and can expect to detect about 22 tax evasion activities per 100 posts (464 out of 2081). With our method, our model will identify the suspicious posts first, then tax officers will manually confirm whether these posts relate to tax evasion. We can expect to identify about 72 real tax evasion activities out of 100 recommended suspicious posts. Therefore, with the same amount of effort, the efficiency can be improved by more than 3 times.

In our current labelled dataset, about 10 percent of the samples have video clips instead of images. Since our current method could not take video clips as input, we replaced the video clips with random noise images for visual feature extraction. Features extracted from random noise images do not provide useful information for our detection task, thereby may compromise the performance of the model. As shown in Figure 3, a marked drop in precision was observed in the image model's performance. For future work, we will implement a video modality module to enhance the model's applicability and improve the model's performance. Another interesting finding of the results is that textural features tend to have higher recall, whereas image features have higher precision, and the combined features can outperform any individual type of features. It would be worthy of further study to investigate the complementary nature of different features and to understand the relationship between them, e.g., what words and images would have higher weights in detecting tax evasion activities or in other applications. 

\section{Conclusions and Roadmap}

In conclusion, we developed a Regtech tool that automatically detects transaction-based tax evasion activities on social media platforms. In the proof-of-concept stage, we collected a dataset of Instagram posts about \#lipstick and manually annotated sampled posts with multiple labels related to sales and tax evasion activities. The dataset provides a solid baseline to understand sales and hidden economy activities on Instagram. We then developed a multi-modal DNN model to automatically detect transaction-based tax evasion activities from the posts. We adopted a modularized structure for the DNN model so that the processing sub-module for each modality can be changed, and new modality could be incorporated easily. Evaluation results confirm the efficiency and effectiveness of the Regtech tool. This tool could help tax authorities to identify audit targets in an efficient and effective manner, and combat social e-commerce tax evasion in scale.

As a roadmap to a full-fledged tool, we aim to extend the detection capability to more products (other than lipstick), and eventually have a robust detection model even if the product is unseen by the model. This continuous process will also expand on our novel dataset. We will make available this dataset to the research community once we develop a prototype to the full-fledged tool. Moreover, we plan to incorporate additional modality to further improve the detection rate (e.g., the tax evasion nature of 58 out of the 2,081 posts can only be decided by a combination of post content and the poster’s information). 

This proof of concept model has attracted attention from both the State Administration of Taxation in the People’s Republic of China (the PRC) and the Australian Federal Police (AFP). Our team has formed a collaborative relationship with the Tax Science and Research Institute, a department under the State Administration of Taxation in the PRC. Collaborating with their frontier research lab, we aim to test this project as part of the designing stage in China’s Golden Tax Project, with the plan to incorporate other products that are popular with Daigous in our model design. For example, due to the COVID-19, there is worldwide shortage of medical protective equipment/masks, the PRC government is keen to regulate the unregistered sales of those products. We are waiting for them to provide a list of these popular products. 

As for the collaboration with AFP, the detection of hidden economy transaction on social media platforms covers several legal aspects, and the AFP is also interested in the regulation of other legal issues such as infringement of IP, fake products and smuggling.  

\begin{acks}
We acknowledge our data and labeling team: Alex Huang, Pei-Wen Sophie Zhong, and Jun Zhao. Special thanks to Fujitsu Australia Limited for providing the computational resources for this study.
\end{acks}

\bibliographystyle{ACM-Reference-Format}
\bibliography{main}


\end{document}